\documentclass[a4paper]{svproc}
\usepackage{graphicx}
\usepackage{multicol}
\usepackage{footmisc}
\usepackage{amsmath}
\usepackage{amsfonts}
\usepackage{subcaption}
\usepackage{placeins}
\usepackage{algorithm}
\usepackage{algpseudocode}
\usepackage{caption}
\usepackage{array}
\usepackage{tabularray}
\usepackage{xcolor}
\usepackage{dashrule}
\usepackage[numbers]{natbib}
\usepackage{url}
\usepackage{wrapfig}

\newcommand{\SE}{\mathrm{SE}}

\newtheorem{assumption}{Assumption}

\begin{document}
\mainmatter

\title{Visual Foresight With a Local Dynamics Model}
\titlerunning{Visual Foresight With a Local Dynamics Model}

\author{Colin Kohler \and Robert Platt}
\authorrunning{Kohler et al.} 
\tocauthor{Colin Kohler, Robert Platt}

\institute{
  Khoury College of Computer Sciences\\
  Northeastern University\\
  Boston, MA 02115, USA\\
  \email{\{kohler.c, r.platt\}@northeastern.edu}
}

\maketitle

\begin{abstract}
Model-free policy learning has been shown to be capable of learning manipulation
policies which can solve long-time horizon tasks using single-step manipulation primitives.
However, training these policies is a time-consuming process requiring large amounts of
data. We propose the Local Dynamics Model (LDM) which efficiently learns the state-transition
function for these manipulation primitives. By combining the LDM with model-free policy 
learning, we can learn policies which can solve complex manipulation tasks using one-step
lookahead planning. We show that the LDM is both more sample-efficient and outperforms other 
model architectures. When combined with planning, we can outperform other model-based and
model-free policies on several challenging manipulation tasks in simulation. 

\keywords{Spatial Action Space, Visual Dynamics Model, 
          Reinforcement Learning, Robotic Manipulation}
\end{abstract}


\section{Introduction}
\label{sec:introduction}

Real-world robotic manipulation tasks require a robot to execute complex motion plans 
while interacting with numerous objects within cluttered environments. Due to the difficulty in
learning good policies for these tasks, a common approach is to simplify policy learning
by expressing the problem using more abstract (higher level) actions such as
end-to-end collision-free motions combined with some motion primitive such as pick, 
place, push, etc. This is often called the \emph{spatial action space} and is used by several 
authors including~\cite{zeng2018learning,platt2019deictic,wang2020policy,wang2022equivariant}.
By leveraging these open-loop manipulation primitives, model-free policy learning learns faster 
and can find better policies. However, a key challenge with this approach is that a large number
of actions need to be considered at each timestep leading to difficulties in learning within a
large $SE(2)$ workspace or an $SE(3)$ workspace of any size.

Due to these challenges, model-based policy learning presents an attractive alternative 
because it has the potential to improve sample efficiency
\cite{sutton1991dyna, gu2016continuous, kaiser2019modelatari}. 
Applying model-based methods to robotics, however, has been shown
to be difficult and often requires reducing the high-dimensional states provided by 
sensors to low-dimensional latent spaces. While these methods have been successfully
applied to a variety of robotic manipulation tasks \cite{tassa2018deepcontrol, lenz2015deepmpc}
they also require a large  amount of training data (on the order of 10,000 to 100,000 examples).

This paper proposes the Local Dynamics Model (LDM) which learns the state-transition
function for the pick and place primitives within the spatial action space. Unlike
previous work which learns a dynamics model in latent space, LDM exploits the
encoding of actions into image-space native to the spatial action space to instead 
learn an image-to-image transition function. Within this image space, we leverage both
the localized effect of pick-and-place actions and the spatial equivariance property
of top-down manipulation to dramatically improve the sample efficiency of our method.
Due to this efficiency, the dynamics model quickly learns useful predictions allowing
us to perform policy learning with a dynamics model which is trained from scratch
alongside the policy. We demonstrate this through our use of a one-step lookahead 
planner which uses the state value function in combination with the LDM to solve many 
different complex manipulation tasks in simulation. 

We make the following contributions. First, we propose the Local Dynamics Model, a novel 
approach to efficiently modelling environmental dynamics by restructuring the transition
function. Second, we introduce a method which leverages the LDM to solve challenging 
manipulation tasks. Our experimental results show that our method outperforms
other model-based and model-free. Our code is available at
\url{https://github.com/ColinKohler/LocalDynamicsModel}.

\section{Related Work}
\label{sec:related}

\paragraph{Robotic Manipulation:}
Broadly speaking, there are two common approaches to learning manipulation policies:
open-loop control and close-loop control. In closed-loop control, the agent controls 
the delta pose of the end-effector enabling fine-tune control of the manipulator. This 
end-to-end approach has been shown to be advantageous when examining contact-rich domains
\cite{levine2016learning}, \cite{jang2017endtoend}, \cite{kalashnikov2018qtopt}. In contrast,
agents in open-loop control apply predefined action primitives, such as pick, place, or push, 
to specified poses within the workspace. This tends to provide more data-efficient learning
but comes at the cost of less powerful policies \cite{mahler2019learning}.

\paragraph{Spatial Action Space}
The spatial action space is an open-loop control approach to policy learning for robotic 
manipulation. Within this domain, it is common to combine planar manipulation with
a fully-convolutional neural network (FCN) which is used as a grasp quality metric
\cite{mahler2019learning} or, more generally, a action-value metric 
\cite{zeng2018learning, wang2020policy}.
This approach has been adapted to a number of different manipulation tasks covering a
variety of action primitives \cite{berscheid2019robot, liang2019knowledge, wu2020spatial}. 

\paragraph{Dynamics Modelling:} Model-Based RL improves data-efficiency by incorporating a
learned dynamics model into policy training. Model-based RL has been successfully applied
to variety of non-robotic tasks \cite{gu2016continuous, drl_cem, kaiser2019modelatari}
but has seen more mixed success in robotics tasks. While 
model-based RL has been shown to work well in robotics tasks with low-dimensional state-spaces
\cite{tassa2018deepcontrol, lenz2015deepmpc}, the high-dimensionality of visual 
state-spaces more commonly seen in robotic manipulation tends to harm performance. 
More modern approaches learn a mapping from image-space to some underlying latent space 
and learn a dynamics model which learns the transition function between these latent states
\cite{minderer2019unsupervised, kossen2019structured}.

More recent work has examined image-to-image dynamics models similar to the video prediction
models in computer vision \cite{finn2016deepvisualforesight}. However, these works typically 
deal with short-time horizon physics such as poking objects \cite{agrawal2016learning} or 
throwing objects \cite{zeng2020tossingbot}. \citeauthor{paxton2019visual} 
\cite{paxton2019visual} and \citeauthor{hoque2020visuospatial} \cite{hoque2020visuospatial}
learn visual dynamics models for pick and place primitives but require a large amount of 
data and time to learn an accurate model. Our work is most closely related to  
\cite{berscheid2021learning} and \cite{wu2022transporters}. In \cite{berscheid2019robot},
\citeauthor{berscheid2021learning}, learn a visual transition model using a GAN architecture 
but only learn pick and push primitives while still requiring a large amount of data. 
Additionally, they only examine a simple bin picking task in their experiments.
\citeauthor{wu2022transporters} \cite{wu2022transporters} learn a visual foresight model
tailored to a suction cup gripper and use it to solve various block rearrangement tasks.
In contrast, we achieve similar sample efficiency using a more complicated parallel jaw
gripper across a much more diverse set of objects and tasks.


\section{Problem Statement}
\label{sec:problem_statement}
\begin{wrapfigure}[16]{r}{0.45\textwidth}
    \vspace{-1cm}
    \centering
    \includegraphics[width=0.45\textwidth]{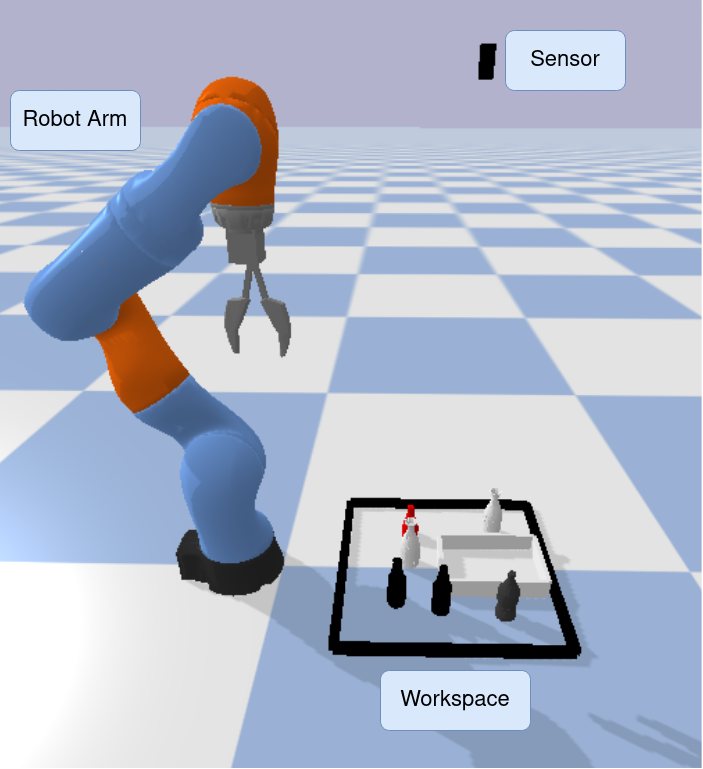}
    \caption{\textbf{The Manipulation Scene}}
    \label{fig:manip_scene}
\end{wrapfigure}

\paragraph{Manipulation as an MDP in a spatial action space:}

\begin{figure}[t]
    \centering
    \includegraphics[width=0.95\textwidth]{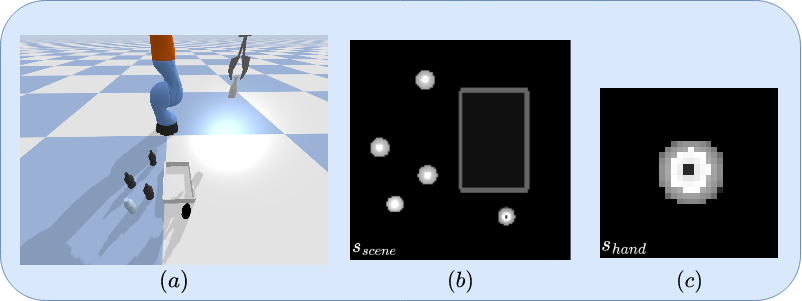}
    \caption{\textbf{MDP State}. (a) The manipulation scene. (b) The top-down image of the workspace,
             $s_{scene}$. (c) The in-hand image, $s_{hand}$.}
    \label{fig:manip_details}
\end{figure}

This paper focuses on robotic manipulation problems expressed as a Markov decision process
in a spatial action space, $\mathcal{M} = (S, A, T, R, \gamma)$, where state is a top down 
image of the workspace paired with an image of the object currently held in the gripper 
(the \emph{in-hand image}) and action is a subset of $SE(2)$. Specifically, state is a pair
of $c$-channel images, $s = (s_{scene}, s_{hand}) \in S_{scene} \times S_{hand}$, where
$s_{scene} \in S_{scene} \subseteq \mathbb{R}^{c \times h \times w}$ is a $c \times h \times w$
image of the scene and $s_{hand} \in S_{hand} \subseteq \mathbb{R}^{c \times d \times d}$ 
is a $c \times d \times d$ image patch that describes the contents of the hand (Figure
\ref{fig:manip_details}). At each 
time step, $s_{scene}$ is set equal to a newly acquired top-down image of the scene. 
$s_{hand}$ is set to the oriented $d \times d$ image patch corresponding to the pose of
the last successful pick. If no successful pick has occurred or the hand is empty, then 
$s_{hand}$ is set to be the zero image. Action $a \in A \subseteq SE(2)$ 
is a target pose for an end effector motion to be performed at the current timestep. 
If action $a$ executes when the hand is holding an object (when the in-hand image is not zero),
then $a$ is interpreted as a place action, i.e. move and then open the fingers. Otherwise,
$a$ is interpreted as a pick, i.e. move and close the fingers. Here, 
$A = A_{pos} \times S^1 \subseteq SE(2)$ spans the robot workspace and 
$A_{pos} \subseteq \mathbb{R}^2$ denotes the position component of that workspace. State
and action are related to each other in that each action corresponds to the pixel in the
state that is beneath the end effector target pose specified by the action.
We assume we have access to a function $h: A \rightarrow \mathbb{Z}^2$ that maps an action
to the pixel corresponding to its position component.

\paragraph{Assumptions:} 

The following assumptions can simplify policy learning and are often reasonable in robotics
settings. First, we assume that we can model transitions with a deterministic function. 
While manipulation domains can be stochastic, we note that \emph{high value} transitions
are often nearly deterministic, e.g. a high value place action often leads to a desired next
state nearly deterministicly. As a result, planning with a deterministic model is often 
reasonable. 

\begin{assumption}[Deterministic Transitions]
\label{ass:deterministic_transitions}
The transition function is deterministic and can therefore be modeled by the function
$s' = f(s,a)$, i.e. the dynamics model.
\end{assumption}

The second assumption concerns symmetry with respect to translations and rotations of
states and actions. Given a transformation $g \in SE(2)$, $g(s)$ denotes the 
state $s = (s_{scene},s_{hand})$ where $s_{scene}$ has been rotated and translated by 
$g$ and $s_{hand}$ is unchanged. Similarly, $g(a)$ denotes the action $a$ rotated and
translated by $g$. 

\begin{assumption}[$SE(2)$ Symmetric Transitions]
\label{ass:symmetric_transitions}
The transition function is invariant to translations and rotations. That is, for any 
translation and rotation $g \in SE(2)$, $T(s,a,s') = T(g(s),g(a),g(s'))$ for all
$s, a, s' \in S \times A \times S$.
\end{assumption}

The last assumption concerns the effect of an action on state. Let $R \subseteq A_{pos}$
be a region of $\mathbb{R}^2$. Given a state $s = (s_{scene},s_{hand})$, let 
$s'_{scene} = \textsc{mask}(s,R) \in S_{scene}$ denote the scene image that is equal to
$s_{scene}$ except that all pixels \emph{inside} $R$ have been masked to zero. In 
the following, we will be exclusively interested in image masks involving the region 
$B_a$, defined as follows:

\begin{definition}[Local Region]
For an action $a = (a_{pos},a_\theta) \in \SE(2)$, let $B_a \subseteq A_{pos}$ denote
the square region with a fixed side length $d$ (a hyperparameter) that is centered at
$a_{pos}$ and oriented by $a_\theta$. 
\end{definition}

We are now able to state the final assumption:

\begin{assumption}[Local Effects]
\label{ass:local_effect}
An action $a \in A$ does not affect parts of the scene outside of $B_a$. 
That is, given any transition $s' = f(s,a)$, it is the case that 
$\textsc{mask}(s,B_a) = \textsc{mask}(s', B_a)$.
\end{assumption}

The bottle arrangement task (Figures \ref{fig:manip_scene}, \ref{fig:manip_details}) is 
an example of a robotic manipulation domain that satisfies the assumptions above. First, 
notice that high value actions in this domain lead to deterministic pick and place outcomes,
i.e. picking up the bottle and placing it with a low probability of knocking it over.
Second, notice that transitions are rotationally and translationally symmetric in 
this problem. Finally, notice that interactions between the hand and the world have 
local effects. If the hand grasps or knocks over a bottle, that interaction typically 
affects only objects nearby the interaction.

\section{Method}
\label{sec:method}
In this section, we first introduce the Local Dynamics Model (LDM) detailing its properties
and model architecture. We then discuss how we combine the LDM with an action
proposal method to perform policy learning through one-step lookahead planning.

\subsection{Structuring the Transition Model}

We simplify the problem of learning the transition function $f : S \times A \rightarrow S$
by encoding Assumptions~\ref{ass:symmetric_transitions} and~\ref{ass:local_effect} as 
constraints on the model as follows. First, given a state $s = (s_{scene},s_{hand})$, we 
partition the scene image $s_{scene}$ into a region that is invariant under $a$, 
$\check{s}_a = \textsc{mask}(s,B_a)$, and a region that changes under $a$, 
$\hat{s}_a = \textsc{crop}(s,B_a)$. Here, 
$\textsc{crop}(s,R) \in \mathbb{R}^{c \times d \times d}$
denotes the $c$-channel $d \times d$ image patch cropped from $s_{scene}$ corresponding to region
$R \subseteq A_{pos}$, resized to a $d \times d$ image. Using this 
notation, we can reconstruct the original scene image by combining 
$\hat{s}_a$ and $\check{s}_a$: 
\begin{equation}
s_{scene} = \textsc{insert}(\hat{s}_a,B_a) + \check{s}_a, 
\end{equation}
where $\hat{s}_a = \textsc{crop}(s,B_a)$ and $\textsc{insert}(\hat{s}_a,B_a)$ inserts the 
crop into region $B_a$ and sets the pixels outside $B_a$ to zero.

\subsection{Local Dynamics Model}
\label{sec:method:dynamics_arch}

\begin{figure}[t]
    \centering
    \includegraphics[width=0.95\textwidth]{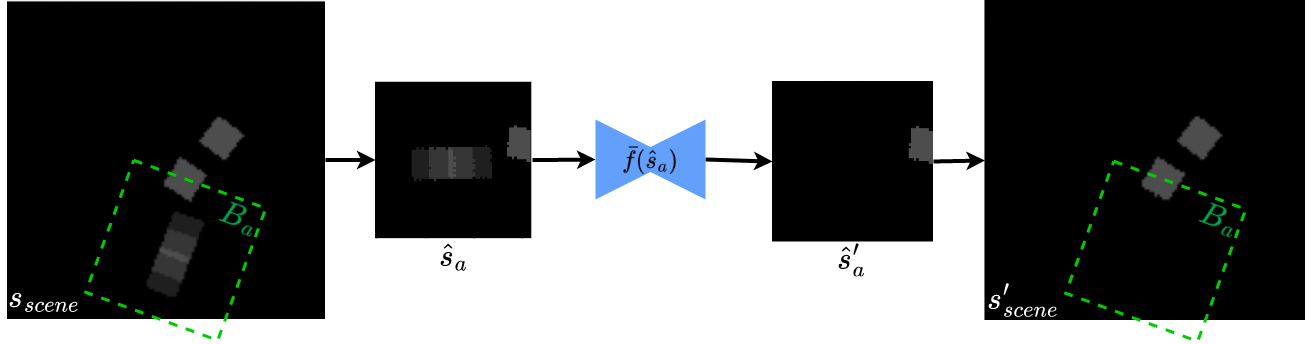} 
    \vspace{1em}
    \rule{8cm}{1pt}
    \includegraphics[width=0.95\textwidth]{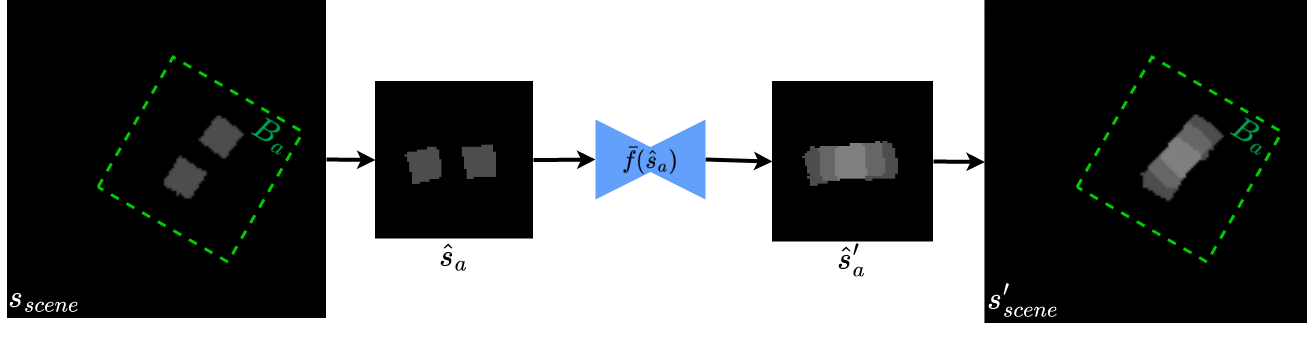}
    \caption{\textbf{Local dynamics model}. In order to predict the next scene image $s'_{scene}$,
    we learn a model $\bar{f}$ that predicts how the scene will change within $B_a$, a neighborhood
    around action $a$. The output of this model is inserted into the original scene image.}
    \label{fig:dynamics_representation}
\end{figure}

Instead of learning $f$ directly, we will learn a function 
$\bar{f} : \mathbb{R}^{c \times d \times d} \rightarrow \mathbb{R}^{c \times d \times d}$ 
that maps the image patch $\hat{s}_a$ onto a new patch $\hat{s}'_a$. Whereas $f$ models 
the dynamics of the entire scene, $\bar{f}$ only models changes in the scene within the 
local region $B_a$. We refer to $\bar{f}$ as the \emph{local dynamics model} (LDM). Given such 
a model, we can define a function $f_{scene}$ as:
\begin{equation}
f_{scene}(s,a) = \textsc{insert}(\bar{f}(\hat{s}_a),B_a) + \check{s}_a,
\label{eqn:local_dynamics}
\end{equation}
where $\hat{s}_a = \textsc{crop}(s,B_a)$. We can reconstruct $f$ as 
$f(s,a) = (f_{scene}(s,a), s'_{hand})$ where $s'_{hand}$ denotes the in-hand image obtained
using the rules described in Section~\ref{sec:problem_statement}. Figure 
\ref{fig:dynamics_representation} illustrates this process for picking and placing in a 
block arrangement task.

Notice that the model in Equation~\ref{eqn:local_dynamics}, $f_{scene}$, satisfies both
Assumptions~\ref{ass:symmetric_transitions} and~\ref{ass:local_effect}. The fact that it 
satisfies Assumption~\ref{ass:local_effect} is easy to see as the local 
dynamics model $\bar{f}$ only models changes in the scene within the local region $B_a$.
It also satisfies Assumption~\ref{ass:symmetric_transitions} because $\hat{s}_a$ is 
invariant under transformations $g \in \SE(2)$ of $s$ and $a$:
\begin{align*}
    \hat{s}_a &= \textsc{crop}(s,B_a) \\
              &= \textsc{crop}(g(s),B_{g(a)}),
\end{align*}
where $g(s)$ rotates state $s$ and $g(a)$ rotates action $a$. As a result,
Equation~\ref{eqn:local_dynamics} is constrained to be equivariant in the
sense that $g(f_{scene}(s,a)) = f_{scene}(g(s),g(a))$.

\begin{figure}[t]
    \centering
    \includegraphics[width=0.95\textwidth]{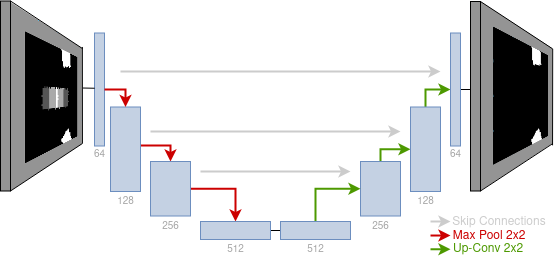} 
    \caption{\textbf{LDM architecture}. Model architecture used in $\bar{f}$, the local 
             dynamics model. Each blue box represents a 3x3 ResNet Block.}
    \label{fig:ldm_model_sm}
\end{figure}

\paragraph{Model Architecture:}

We model the local dynamics model, $\bar{f}$, using the UNet model architecture 
shown in Figure~\ref{fig:ldm_model_sm} with four convolution and four deconvolution layers. 
It takes as input the image patch 
$\hat{s}_a \in \mathbb{R}^{c \times d \times d}$, 
and outputs a patch from the predicted next state, 
$\hat{s}'_a \in \mathbb{R}^{c \times d \times d}$. 
The size of this image patch must be large enough to capture the effects of pick and
place actions, but small enough to ignore objects not affected by the current interaction. 
In our experiments, we set $d=64$ pixels which corresponds to roughly $20cm$ in the workspace.

\paragraph{Loss Function:}

$\bar{f}$ is trained using a reconstruction loss, i.e. a loss which measures the difference
between a predicted new state image patch and a ground truth image patch. Typically, this is
accomplished using a pixel-wise L2 loss~\cite{hinton2006reducing}. However, we instead model
pixel values as a multinomial probability distribution over 21 different possible values for 
each pixel (in our case, these are depth values since we use depth images). This enables us
to use a cross entropy loss, which has been shown to have better performance relative to an L2
loss~\cite{van2016pixel}. We were able to improve performance even further by using a focal
loss rather than a vanilla cross entropy loss~\cite{Lin_2017_focal}. This alleviates the 
large class imbalance issues that arise from most pixels in $\hat{s}_a$ having the same value
and focuses learning on parts of the pixel space with the most challenging dynamics.

\subsection{Policy Learning}
\label{sec:policy_learning}

While there are a variety of ways to improve policy learning using a dynamics model, here 
we take a relatively simple one-step lookahead approach. We learn the state value function
$V_\psi(s)$, and use it in combination with the dynamics model to estimate the $Q$ function,
$\hat{Q}(s,a) = V_\psi(f(s,a))$. A key challenge here is that it is expensive 
to evaluate $\max_{a \in A} \hat{Q}(s,a)$ or $\arg\max_{a \in A} \hat{Q}(s,a)$ over large 
action spaces (such as the spatial action space) because 
the forward model must be queried separately for each action. We combat this problem by 
learning an approximate $Q$ function that is computationally cheap to query and use it to
reduce the set of actions over which we maximize. Specifically, we learn a function $Q_\theta$
using model-free $Q$-learning: 
$Q_\theta(s,a) \leftarrow r + \gamma \max_{a' \in A} Q_\theta(s',a')$. Then, we define a policy
$\pi_\theta(a | s) = \sigma_A(Q_\theta(s,a))$, where $\sigma_A$ denotes the softmax function
over the action space $A$ with an implicit temperature parameter $\tau$. We sample a small set
of high quality actions $\bar{A}_N \subseteq A$ by drawing $N$ action samples from
$\pi_\theta(a | s)$. Now, we can approximate 
$\max_{a \in A} \hat{Q}(s,a) \approx \max_{a \in \bar{A}_N} \hat{Q}(s,a)$. The target for 
learning $V_\psi$ is now $V_\psi(s) \leftarrow r + \max_{a \in \bar{A}_N} \hat{Q}(s,a)$. The 
policy under which our agent acts is $\pi(a | s) = \sigma_{\bar{A}_N}(\hat{Q}(s,a))$. We 
schedule exploration by decreasing the softmax temperature parameter over the course of 
learning.

We model $Q_\theta$ using a fully-convolutional neural network which takes as input the
top-down heightmap $s_{scene}$ and outputs a 2-channel action-value map 
$(Q_{pick}, Q_{place}) \in \mathbb{R}^{2 \times r \times h \times w}$
where $Q_{pick}$ correlates with picking success and $Q_{place}$ to placing success.
The orientation of the action is represented by discretizing the space of $SO(2)$ 
rotations into $r$ values and rotating $s$ by each $\theta$ value. $V_\psi$ is modeled 
as standard convolutional neural network which takes the state $s$ as input and 
outputs the value of that state. We use two \textit{target networks} parameterized by 
$\theta^-$ and $\psi^-$ which are updated to the current weights $\theta$ and $\psi$ 
every $t$ steps to stabilize training. 

\subsection{Sampling Diverse Actions}
\label{sec:action_sampling}

When evaluating 
$\max_{a \in \bar{A}_N} \hat{Q}(s,a)$ and $\pi(a | s) = \sigma_{\bar{A}_N}(\hat{Q}(s,a))$, 
it is important to sample a diverse set of actions $\bar{A}_N$. The problem is that
$\sigma(Q_\theta,\cdot))$ can sometimes be a low entropy probability distribution with a small 
number of high-liklihood peaks. If we draw $N$ independent samples directly from this 
distribution, we are likely to obtain multiple near-duplicate samples. This is unhelpful since
we only need one sample from each mode in order to evaluate it using $V_{\psi}(f(s,a))$.
A simple solution would be to sample \emph{without} replacement. Unfortunately, as these peaks 
can include a number of actions, we would have to draw a large number of samples in order
to ensure this diversity.
To address this problem, we use an inhibition technique similar
to non-maximum suppression where we reduce the distribution from which future samples are drawn
in a small region around each previously drawn sample. Specifically, we draw a sequence of 
samples, $a_1, \dots, a_N$. The first sample is drawn from the unmodified distribution
$Q_\theta(s,\cdot)$. Each successive sample $j \neq N$ is drawn from a distribution 
$Q_\theta(s,\cdot) - \beta \sum_{i=1}^j \mathcal{N}(a_i,\sigma^2)$, where $\mathcal{N}$ 
denotes the standard normal distribution in $\mathbb{R}^3$, and $\beta$ and $\sigma^2$ are 
constants. Here, we have approximated $\SE(2)$ as a vector space $\mathbb{R}^3$ in order to 
apply the Gaussian. Over the course of training, we slowly reduce $\beta$ as the optimal policy
is learned.


\section{Experiments}
\label{sec:experiments}

We performed a series of experiments to test our method. First, we investigate the 
effectiveness of the Local Dynamics Model (LDM) by training the model in isolation on
pre-generated offline data. Second, we demonstrate that we can learn effective 
policies across a number of complex robotic manipulation tasks. 

\paragraph{Network Architecture:}
A classification UNet with bottleneck Resnet blocks \cite{resnet} is used as the
architecture of the LDM. A similar network architecture is used for 
the Q-value model, $Q_\theta$, with the exception of using basic Resnet blocks.
The state value model, $V_\psi$, is a simple CNN with basic Resnet blocks and
two fully-connected layers. The exact details for the number of layers and hidden units
can be found in our Github repository.

\paragraph{Implementation Details:}
The workspace has a size of $0.4m \times 0.4m$ and $s_{scene}$ covers the workspace with 
a heightmap of size of $128 \times 128$ pixels. We use $8$ discrete rotations equally
spaced from $0$ to $\pi$. 
The target network is synchronized every 100 steps. We used the Adam optimizer
\cite{kingma2014adam}, and the best learning rate and its decay were chosen to be
$10^{-3}$ and $0.95$ respectively. The learning rate is multiplied by the decay every 
$2000$ steps. We use the prioritized replay buffer \cite{per} with prioritized replay 
exponent $\alpha = 0.6$ and prioritized importance sampling exponent $B_0 = 0.0$ 
annealed to $1$ over training. The expert transitions are given a priority bonus of 
$\epsilon_d = 1$ as in \citeauthor{hester} \cite{hester}. The buffer has a size of
$10000$ episodes. Our implementation is based on PyTorch \cite{paszke2019pytorch}.

\begin{figure}
    \centering
    \begin{subfigure}{0.24\columnwidth}
        \centering
        \includegraphics[width=0.95\linewidth]{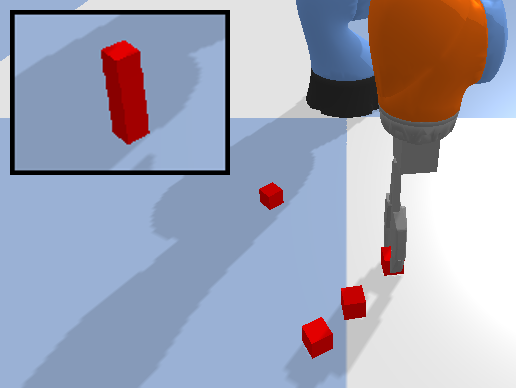}
        \caption{Block Stacking}
        \label{fig:block_stacking_ex}
    \end{subfigure}
    \begin{subfigure}{0.24\columnwidth}
        \centering
        \includegraphics[width=0.95\linewidth]{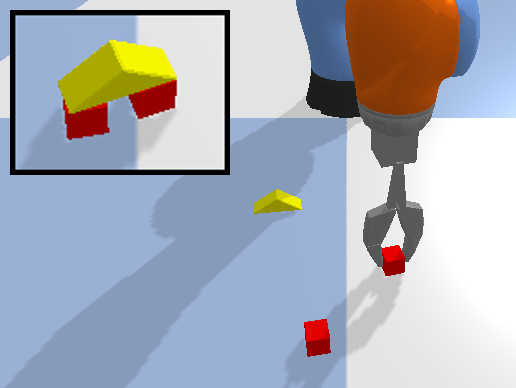}
        \caption{House Building}
        \label{fig:house_building_ex}
    \end{subfigure}
    \begin{subfigure}{0.24\columnwidth}
        \centering
        \includegraphics[width=0.95\linewidth]{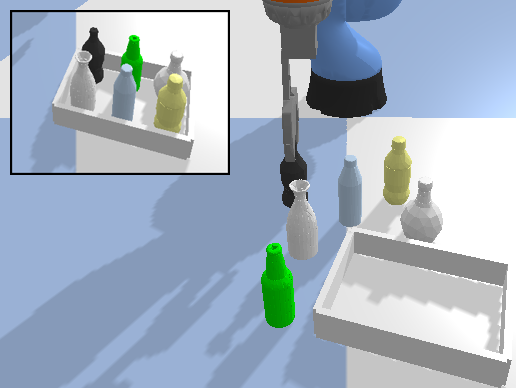}
        \caption{Bottle Arrange}
        \label{fig:bottle_arrangement_ex}
    \end{subfigure}
    \begin{subfigure}{0.24\columnwidth}
        \centering
        \includegraphics[width=0.95\linewidth]{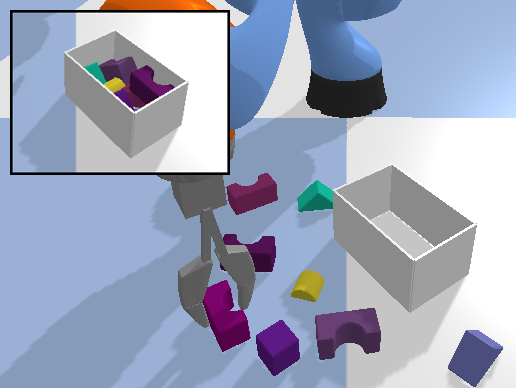}
        \caption{Bin Packing}
        \label{fig:bin_packing_ex}
    \end{subfigure}
    \caption{\textbf{Tasks}. The window in the top-left corner shows the
             goal state.}
    \label{fig:domain_ex}
\end{figure}

\paragraph{Task Descriptions:} 
For all experiments, both the training and testing is preformed in the PyBullet 
simulator \cite{pybullet}. In the block stacking domain, three cubes are placed
randomly within the workspace and the agent is tasked with placing these blocks
into a stable stack. In the house building domain, two cubes and one triangle are
placed randomly within the workspace and the agent is tasked with placing the 
triangle on top of the two cube blocks. In the bottle arrangement domain, the 
agent needs to gather six bottles in a tray. These three environments have spare 
rewards ($+1$ at goal and $0$ otherwise).

In the bin packing domain, the agent must compactly pack eight blocks into a bin while 
minimizing the height of the pack. This environment uses a richer reward function and 
provides a positive reward with magnitude inversely proportional to the highest point 
in the pile after packing all objects. Example initial and goal configurations for these 
domains can be seen in Figure \ref{fig:domain_ex}.

\subsection{Accuracy of the Local Dynamics Model}
\label{sec:experiments:forward_model}

\begin{table}
    \centering
    \begin{tblr}{
    hline{1,3,7,9,12} = {1pt},
    cell{1}{2} = {c = 2}{halign=c},
    cell{1}{5} = {c = 2}{halign=c},
    cell{7}{2} = {c = 2}{halign=c},
    cell{7}{5} = {c = 2}{halign=c},
    row{2} = {halign=c},
    columns = {halign=c},
    stretch=0,
    }
    & Block Stacking & & & House Building & \\
    \cline{2-3, 5-6}
    Method & L1 & SR & & L1 & SR \\
    Naive & $30.3 \pm 1.7$ & $38 \pm 5.6$ & & 
            $30.4 \pm 1.5$ & $39.1 \pm 1.9$ \\
    LDM(128) & $14.5 \pm 2.2$ & $70 \pm 1.8$ & & 
               $10.9 \pm 0.24$ & $70.7 \pm 0.6$ \\
    LDM(64) & $\mathbf{8.76 \pm 0.1}$ & $\mathbf{83.4 \pm 0.6}$ & & 
              $\mathbf{5.88 \pm 0.2}$ & $\mathbf{77.9 \pm 1.1}$ \\
    \\
    & Bottle Arrangement & & & Bin Packing & \\
    \cline{2-3, 5-6}
    Method & L1 & SR & & L1 & SR \\
    Naive & $48.9 \pm 0.9$ & $43.8 \pm 4.4$ & & 
            $77.2 \pm 0.71$ & $35.4 \pm 0.8$ \\
    LDM(128) & $43.6 \pm 0.79$ & $58.6 \pm 1.3$ & & 
               $93.3 \pm 1.9$ & $60.4 \pm  2.1$ \\
    LDM(64) & $\mathbf{32.5 \pm 1.8}$ & $\mathbf{66 \pm 1.9}$ & & 
              $\mathbf{48.8 \pm 0.9}$ & $\mathbf{65.9 \pm 0.5}$ \\
    & 
    \end{tblr}
    \vspace{1em}
    \caption{\textbf{Dynamics Model Performance}. Final performance for the $4$ domains on 
             the different dynamics models. The results show the mean and standard deviation
             averaged over 3 random seeds.
             L1 denotes the L1-pixelwise difference between the predicted observation and 
             the true observation. Lower is better. SR denotes the success rate (\%) for
             the action. Higher is better.}
    \label{tab:forward_model_benchmark}
\end{table}

\paragraph{Experiment:}

We generate $5$k steps of noisy expert data for each of the domains in Figure 
\ref{fig:domain_ex} by rolling out a hand coded stochastic policy. For the block stacking 
and house building domains we train the models for $5$k iterations of optimization. For the
bottle arrangement and bin packing domains we train the models for $10$k iterations. 

\paragraph{Metrics:}
We examine two metrics of model accuracy: 1.) the 
L1-pixelwise difference between the predicted observation and the true observation and 
2.) the success rate of the action primitives. A pick action is defined as a success 
if the model correctly predicts if the object will be picked up or not. Similarly, a
place action is defined as a success provided the model correctly predicts the pose
of the object after placement. The L1 difference provides a low level 
comparison of the models whereas the success rate provides a higher level view which is 
more important for planning and policy learning.

\paragraph{Baselines:} We compare the performance of three dynamics models.
\begin{enumerate}
    \item \textbf{LDM(64):} Local Dynamics Model with a crop size of $64$ pixels.
    \item \textbf{LDM(128):} Local Dynamics Model with a crop size of $128$ pixels.
    \item \textbf{Naive:} UNet forward model with $128 \times 128$ input and output
          size. The action is encoded by concatenating a binary mask of the action
          position onto the state $s$. 
\end{enumerate}

\begin{wrapfigure}[15]{r}{0.45\textwidth}
    \vspace{-0.5cm}
    \centering
    \includegraphics[width=0.45\textwidth]{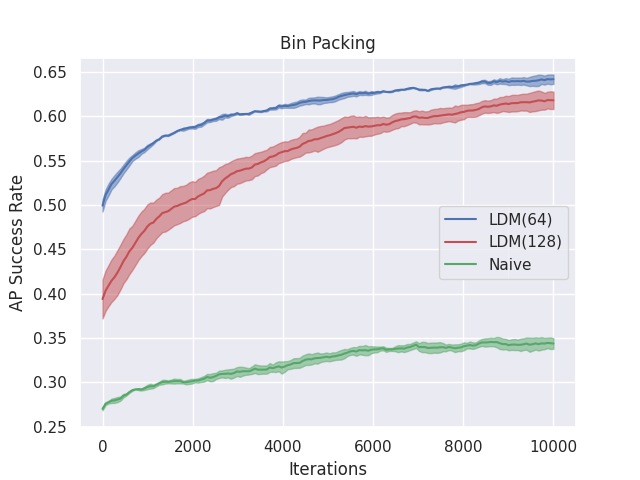}
    \caption{\textbf{Sample Efficiency}. Action primitive success rate for
             bin packing. Results averaged over three random seeds. Shading denotes standard error.}
    \label{fig:bin_packing_forward_model_benchmark}
\end{wrapfigure}

\paragraph{Results:}
In Table \ref{tab:forward_model_benchmark}, we summarize the accuracy of the 
models in the four domains on a held-out test set. While both LDM(64) and LDM(128) are
able to generate realistic images in non-cluttered domains, we find that defining a 
small localized area of affect to be vital in cluttered domains such as bin packing.
The most common failure mode occurs when the model overestimates the stability of object
placements. For example, it has difficulties in determining the inflection point when 
stacking blocks which will lead to the stack falling over. Equally important to the 
final performance of the models is how efficiently they learn.
In Figure \ref{fig:bin_packing_forward_model_benchmark}, the action primitive 
success rate is shown over training for the bin packing environment. The sample 
efficiency of LDM(64) makes it much more useful for policy learning as the faster
the dynamics model learns the faster the policy will learn.

\subsection{Policy Learning}
\label{sec:experiments:policy_learning}

Here, we evaluate our ability to use the local dynamics model to learn policies that solve
the robotic manipulation tasks illustrated in Figure~\ref{fig:domain_ex}. In each of these domains,
the robot must execute a series of pick and place actions in order to arrange a collection of 
objects as specified by the task. These are sparse reward tasks where the agent gets a non-zero 
reward only upon reaching a goal state. As such, we initialize the replay buffer for all agents
with $100$ expert demonstration episodes in order to facilitate exploration. 

\paragraph{Baselines:}
We compare our approach with the following baselines.
\begin{enumerate}
    \item \textbf{FC-DQN:} Model-free policy learning using a fully-convolutional neural network
          to predict the q-values for each action in the spatial-action space. Rotations
          are encoded by rotating the input and output for each $\theta$ 
          \cite{zeng2018learning}.
    \item \textbf{Random Shooing (RS):} RS samples $K$ candidate action sequences from a uniform 
          distribution and evaluates each candidate using the dynamics module. The 
          optimal action sequence is chosen as the one with the highest return 
          \cite{ross2011reduction, richards2005robust}. Due to the size of the action space,
          we restrict action sampling to only sample actions which are nearby or on obejcts 
          within the workspace.
    \item \textbf{Dyna-Q:} FC-DQN model trained Dyna-style where training iterates between two 
          steps. First, data is gathered using the current policy and used to learn the
          dynamics model. Second, the policy is improved using synthetic data generated 
          by the dynamics model. At test time only the policy is used \cite{sutton1991dyna}.
\end{enumerate}

For fairness, all algorithms use the same model architecture. For RS and Dyna-Q, an extra 
head is added onto the state value model after the feature extraction layers in order to 
predict the reward for that state. When a model is not used, such as the value model for RS,
they are not trained during that run. The forward model is not pretrained in any of the algorithms
considered. All algorithms begin training the forward model online using the on-policy data
contained in the replay buffer -- the same data used to train the policy. 

\begin{figure}
    \centering
    \begin{subfigure}{0.45\columnwidth}
        \includegraphics[width=0.95\linewidth]{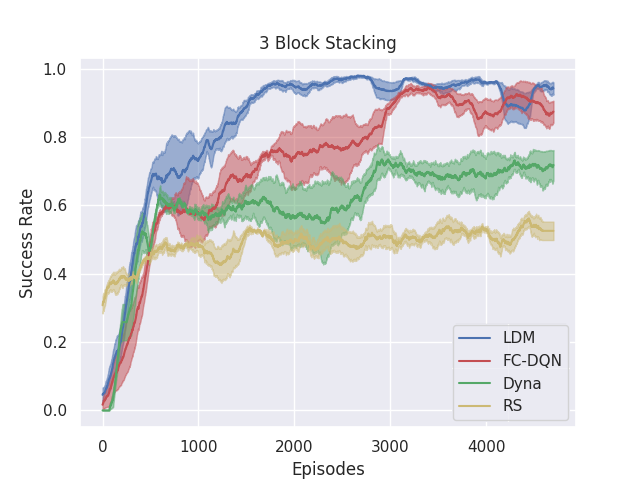}
        \label{fig:block_stacking_3_learning_curve}       
    \end{subfigure}%
    \begin{subfigure}{0.45\columnwidth}
        \includegraphics[width=0.95\linewidth]{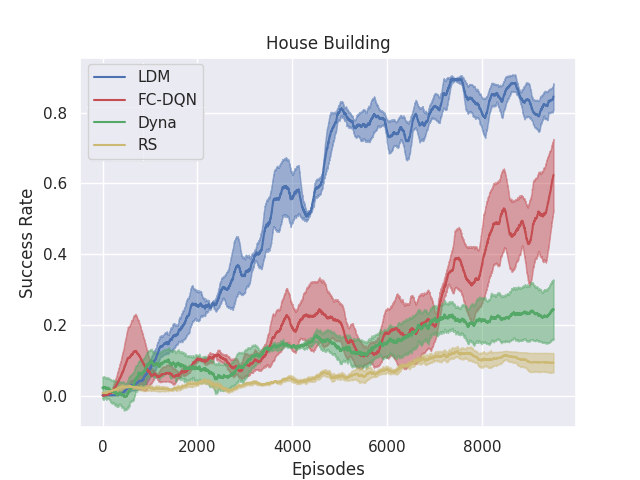}
        \label{fig:house_2_learning_curve}
    \end{subfigure} 
    \begin{subfigure}{0.45\columnwidth}
        \includegraphics[width=0.95\linewidth]{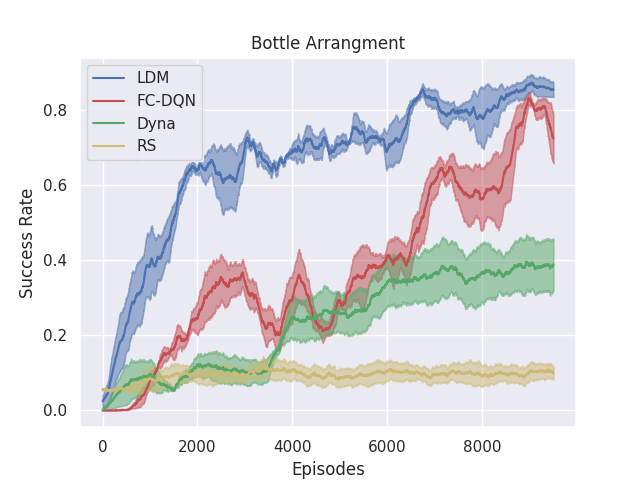}
        \label{fig:bottle_tray_learning_curve}       
    \end{subfigure}%
    \begin{subfigure}{0.45\columnwidth}
        \includegraphics[width=0.95\linewidth]{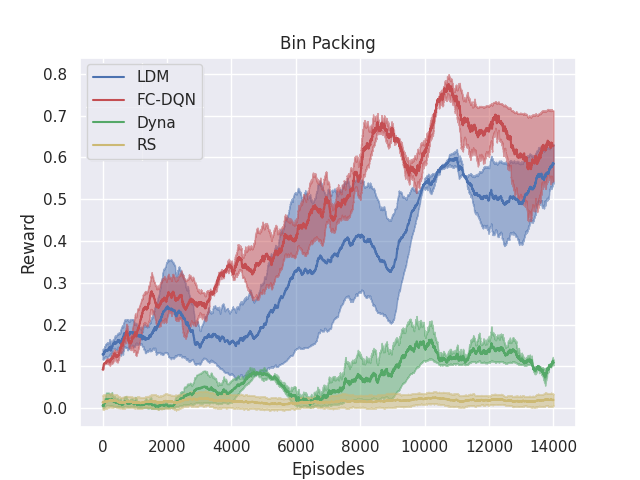}
        \label{fig:bin_packing_learning_curbe}
    \end{subfigure} 
    \caption{\textbf{Simulation Experiment Evaluation}. Evaluation performance of
             greedy-policy. Models are trained until LDM reaches convergence. 
             Results averaged over 3 random seeds. Shading denotes standard error.}
    \label{fig:learning_curves}
\end{figure}

\paragraph{Results:} 

The results are summarized in Figure~\ref{fig:learning_curves}. They show that our method
(shown in blue) is more sample efficient than FC-DQN in all domains except bin packing. 
We attribute the under-performance in bin packing to the difficult transition 
function that the state prediction model must learn due to the varied geometry of the
blocks interacting with each other. LDM significantly out-preforms the model-based 
baselines in all domains. RS preforms poorly even with a high quality state prediction 
model due to the low probability of randomly sampling a good trajectory in large actions 
spaces. Dyna-Q performs similarly poorly due to the minute differences between the simulated 
experiences and the real experiences cause the policy learned to preform worse on 
real data.

\subsection{Generalization}

\begin{table}[t]
    \centering
    \begin{tblr}{
    hline{1,3,6} = {1pt},
    cell{1}{2} = {c = 2}{halign=c},
    cell{1}{5} = {c = 2}{halign=c},
    cell{6}{2} = {c = 2}{halign=c},
    cell{6}{5} = {c = 2}{halign=c},
    row{2} = {halign=c},
    columns = {halign=c},
    stretch=0,
    }
    & Block Stacking & & & Bottle Arrangement \\
    \cline{2-3, 5-6}
    Method & 4 Block & 5 Block & & 5 Bottle & 6 Bottle \\
    RS & 48 & 23 & &  8 & 4 \\
    FC-DQN & 98 & \textbf{89} & & 82 & 48 \\
    LDM & \textbf{99} & 84 & & \textbf{86} & \textbf{65} \\
    \end{tblr}
    \vspace{1em}
    \caption{\textbf{Generalization Experiment}. We show the success rate (\%) of zero-shot
             generalization over 100 episodes. Higher is better.} 
    \label{tab:generalization}
\end{table}

One advantageous property of model-based RL, is its ability to generalize to unseen 
environments provided the underlying dynamics of the environments remains similar. In
order to test how well LDM generalizes, we trained LDM, FC-DQN, and RS on the block 
stacking and bottle arrangement domains on a reduced number of objects and evaluated 
them with an increased number of objects, i.e. zero-shot generalization. Specifically, 
we trained our models on 3 block stacking and evaluated them on 4 and 5 block stacking.
Similarly, we trained our models on 4 bottle arrangement and evaluated them on 5 and 6 
bottle arrangement. As shown in Table \ref{tab:generalization}, LDM is more effective 
for zero-shot generalization when compared to both the model-free (FC-DQN) and model-based
(RS) baselines. 

\section{Limitations and Future Work}
\label{sec:limitations}
This work has several limitations and directions for future research. The most glaring
of these is our use of a single-step lookahead planner for policy learning. One large 
advantage of model-based methods is their ability to plan multiple steps ahead to find
the most optimal solution. For instance in bin packing, our single-step planner will
occasionally greedily select a poor action which results in the final pack being
taller whereas a multi-step planner would be able to avoid this action by examining 
the future consequences. Similarly, model-based methods have been shown to work well
in multi-task learning where a more general model is learned and leveraged across a 
number of tasks. While we show that we can use the LDM for zero-shot generalization, our
planning approach is more tailored to learning single-policies. The LDM on the 
other hand, is shown to be capable of modeling the interactions between many different 
objects across many different tasks making it ideal for use in multi-task learning.

In terms of the LDM, we believe their are two interesting avenues for future work.
First, due to our modeling of the pixels as probability distributions, we can easily
estimate the uncertainty of the LDM's predictions by calculating the pixelwise 
entropy of the model output. This could prove useful when planning by allowing us
to avoid taking actions which the LDM is more uncertain about leading to more robust
solutions. Secondly, although we encode $SE(2)$ equivariance into the LDM by
restructuring the dynamics function, we could also explore the use of equivriatant 
CNNs in the LDM architecture. These equviariant CNNs have been shown to greatly
improve sample efficiency across a wide number of tasks and have recently started 
being applied to robotic manipulation tasks similar to those we present in this work.


\section{Conclusion}
\label{sec:conclusion}

In this paper, we propose the Local Dynamics Model (LDM) approach to forward modeling
which learns the state-transition function for pick and place manipulation primitives.
The LDM is able to efficiently learn the dynamics of many different objects faster and
more accurately compared to similar methods. This sample efficiency is achieved by 
restructuring the transition function to make the LDM invariant to both objects outside
the region near the action and to transformations in $SE(2)$. We show that the LDM can 
be used to solve a number of complex manipulation tasks through the use of a single-step
lookahead planning method. Through the combination of the LDM with our planning method 
which samples a diverse set of actions, our proposed method is able to outperform the 
model-free and model-based baselines examined in this work.

\vspace{3em}
\begingroup
\let\clearpage\relax
\bibliographystyle{abbrvnat}
\bibliography{main}
\endgroup

\end{document}